\documentclass[11pt]{article}
\usepackage[letterpaper,margin=1in]{geometry}
\usepackage{microtype}
\usepackage{amsmath,amssymb,amsthm}
\usepackage{booktabs}
\usepackage{tabularx}
\usepackage{array}
\usepackage{xspace}
\usepackage{graphicx}
\usepackage{tikz}
\usepackage{xcolor}
\usepackage{cite}
\usepackage[hidelinks,hypertexnames=false]{hyperref}
\usepackage{xurl}
\usepackage[section]{placeins}
\usetikzlibrary{arrows.meta,positioning,fit,calc,backgrounds}
\hypersetup{
  hidelinks,
  pdftitle={CARD: A Proof-Carrying Certificate-to-Action Interface for Relative-Measurement Graphs},
  pdfauthor={Yushan Li}
}
\definecolor{cardBlue}{HTML}{1F4E79}
\definecolor{cardTeal}{HTML}{1F7A7A}
\definecolor{cardGold}{HTML}{B57C1B}
\definecolor{cardSlate}{HTML}{5A6673}
\tikzset{cardArrow/.style={-{Latex[length=2.3mm]},thick,draw=black!75},cardBox/.style={draw=cardBlue,rounded corners=2pt,thick,fill=cardBlue!7,align=center,inner sep=4pt},cardEvidence/.style={draw=cardTeal!85!black,rounded corners=2pt,thick,fill=cardTeal!10,align=center,inner sep=4pt},cardAction/.style={draw=cardGold!80!black,rounded corners=2pt,thick,fill=cardGold!16,align=center,inner sep=3pt},cardNote/.style={draw=black!45,rounded corners=2pt,dashed,fill=white,align=left,inner sep=3pt}}
\newcommand{\CARD}{\textsc{card}\xspace}

\newcolumntype{Y}{>{\raggedright\arraybackslash}X}
\newcolumntype{P}[1]{>{\raggedright\arraybackslash}p{#1}}
\theoremstyle{definition}

\newtheorem{theorem}{Theorem}
\newtheorem{corollary}{Corollary}
\newtheorem{proposition}{Proposition}
\title{CARD: A Proof-Carrying Certificate-to-Action Interface for Relative-Measurement Graphs}
\author{Yushan Li}
\date{}
\begin{document}
\maketitle
\begin{abstract}
Geometric certificates make mathematical claims about specified estimator objects, while robotic action requires a physical transform, a service decision, and timely proof support. We present \CARD{}, a proof-carrying interface that keeps these objects distinct and records how they combine. Its lifted-fiber formulation separates semantic motion from witness repair and characterizes when projected and proof-carrying costs coincide. Its actionability result gives an exact useful-service law: uniform sampling without replacement produces a hypergeometric yield, while sampling from the eligible population gives one useful service per request under the observed live-confirmation condition. The primary study contains 404 Full16 cases from 20 sorties of an AA15--X500 airborne survey system. All 404 checker-confirmed optimizer points are reference-discrepant, with median rotation and translation discrepancies of $1.9891^\circ$ and $21.6322\,\mathrm{m}$. The paired causal-navigation transforms are reference-consistent in all 404 cases, with medians of $0.0036566^\circ$ and $0.0104519\,\mathrm{m}$. Across five matched request budgets, the largest difference between the hypergeometric prediction and the observed ordinary-random mean is $0.8412$ services. Ordinary matched random reaches $0/50$ target successes, while actionability-matched random reaches $50/50$ and matches the \CARD{} policy. Public downstream evaluation on 11 KITTI, EuRoC, and TUM SLAM cases measures optimizer wall-time ratios of $1.128\times$--$67.65\times$ for random versus \CARD{} graph interventions. Native solver records locate this difference in elimination fill, Cholesky factor size, and numeric factorization. Together, these results connect certificate semantics, physical execution, proof acquisition, and graph computation through explicit system objects.
\end{abstract}
\section{Introduction}
Checkable geometric estimation has changed what a robotics system can know about an optimizer. A certificate can bound an objective or verify a registration without relying on the optimizer's internal stopping rule. This mathematical statement is valuable, but it does not by itself specify the transform that the robot should execute. Execution also depends on coordinate conventions, causal navigation, service support, proof availability, and the action applied to the downstream graph.

These objects are often compressed into one accept/reject decision. That compression becomes problematic when a checker-confirmed optimizer point and a navigation-supported transform are both valid outputs of their own stages but represent different quantities. The same issue appears over time. A semantic action may remain unchanged while its exact proof witness expires, is replaced, or cannot be acquired before a deadline. A system that records only confidence or acceptance cannot express either distinction.

The problem studied here is the interface between mathematical certification and executable service. The interface must preserve the object certified by the checker, define the physical transform independently, and state when a proof token can support that transform as a service. It must also respect chronology. Request-time decisions may use current geometry and completed past outcomes, while checker results and reference measurements arrive later. This ordering makes proof acquisition a causal resource-allocation problem rather than a post hoc label-selection problem.

\CARD{} addresses this problem with a proof-carrying fork--join contract. Each image/event opportunity defines a frozen Full16 certificate instance and a causal-navigation transform. The support predicate evaluates whether the transform can instantiate the service before a certificate is purchased. If the request completes and the checker confirms the frozen instance, the token joins the unchanged support decision and physical transform to create a proof-backed service. The resulting service record can then drive a graph intervention.

Two theoretical results describe this contract. The lifted-fiber formulation places a semantic action and its exact witness in one state. It decomposes transition cost into semantic movement and witness repair on the same trace, which yields a sharp condition for equality between projected and proof-carrying costs. The actionability result describes acquisition at a fixed request budget. Uniform sampling from a population with $K$ eligible candidates among $N$ total candidates has a hypergeometric useful-service yield. Sampling from the eligible population gives yield $b$ for budget $b\leq K$ under the observed live-confirmation condition.

The empirical study follows these two results. The primary dataset contains 404 non-overlapping Full16 cases from 20 sorties of an AA15--X500 airborne survey system over mountainous, village, and town scenes. The paired evaluation separates certificate-side optimizer coordinates from causal-navigation transforms at the same image/event identities. A 170-candidate policy block then tests the predicted useful-service law at five request budgets. Public KITTI, EuRoC, and TUM SLAM cases evaluate the computational consequence of the recorded graph interventions with native sparse-solver measurements.

The paper contributes an explicit certificate-to-action state model, a lifted-fiber decomposition with a projected/lifted coincidence condition, and an exact request-time actionability law. The experiments provide a 404-case paired certificate/action study, a matched acquisition study with $N=170$ and $K=155$, and an 11-case public downstream analysis that measures elimination fill, factor size, numeric work, and optimizer time.
\section{Related Work}
\subsection{Certified and robust geometric estimation}
Certifiable estimation and synchronization methods provide checkable statements about an optimizer or a registration, including certifiably correct synchronization and robust geometric registration~\cite{rosen2019sesync,singer2011angular,yang2021teaser}. Their outputs concern an estimator, an objective value, or a registration relation. \CARD uses these estimator-side statements as one stage in a larger certificate-to-action contract that also specifies the physical transform, witness lifetime, and service obligations.

The distinction is also useful for interpreting the certificate-side estimate. A certificate can establish that a particular correspondence problem has a certified solution while leaving its coordinate representation subject to correspondence geometry, timing, or camera-specific conventions. The physical transform in our system is consequently computed by causal navigation and evaluated against the processed navigation reference. This division preserves the mathematical certificate while assigning physical meaning to a separately specified output.
\subsection{Acquisition, selective service, and verification}
Selective prediction and reject-option learning study how a system trades coverage against the risk of making an unsupported prediction~\cite{chow1970reject,elyaniv2010selective,cortes2016rejection,geifman2019selectivenet}. Learning-to-defer makes the handoff itself part of the decision interface~\cite{madras2018defer}. Loop-closure verification and trust-but-verify systems instead decide whether a proposed hypothesis should be accepted, rejected, or deferred~\cite{tian2016multimodel,burnett2023tbv}. These literatures clarify the value of explicit refusal and verification states, but their controlled objects differ from a proof token whose support is reused across a chronological service stream.

The acquisition problem in \CARD assigns every request an identity, cost, latency, and potential accepted token. A token can support several component/proposal obligations and can expire or be invalidated. The selected request set controls both the information purchased and the exact witness fibers available to a later service. \CARD{} combines these token-level semantics with explicit refusal and verification states.
\subsection{Graph intervention and sparse computation}
Robust graph estimation manipulates measurement weights, latent switches, or coherent subsets to control the influence of unreliable constraints~\cite{sunderhauf2012switchable,agarwal2013dcs,olson2013mixtures,latif2013robust,yang2020gnc}. CARD acts at a different layer: it selects proof-carrying services whose graph interventions are evaluated under matched budgets. Graph-based SLAM and factor-graph estimation provide the geometric substrate for this downstream study~\cite{kuemmerle2011g2o,grisetti2010tutorial,dellaert2017factorgraphs,cadena2016slam}. These references motivate the pose-graph representation and nonlinear optimization. The present study measures how selected intervention topology changes the cost of the same graph problem.

Sparse direct solvers make topology computationally consequential because symbolic elimination determines fill and factor size before numerical values are processed~\cite{amestoy1996amd,rose1976vertexelimination}. CARD measures these native symbolic and numeric stages directly.

\begin{table}[!b]
\centering
\caption{Scientific objects and outputs related to the certificate-to-action contract.}
\label{tab:related-work-objects}
\footnotesize
\setlength{\tabcolsep}{3pt}
\begin{tabularx}{\textwidth}{@{}l Y Y Y@{}}
\toprule
Literature line & Controlled or certified object & Typical output & Role in the \CARD{} interface \\
\midrule
Certifiable estimation and synchronization & Estimator, registration, or synchronization objective & Certificate, optimizer guarantee, or robust registration & Mathematical evidence for the certificate channel and its frozen objective \\
Robust graph estimation & Measurement weight, latent switch, or coherent subset & Weights, accepted subset, or robust graph estimate & Graph-side mechanisms applied after service selection \\
Selective prediction and defer & Example-level prediction risk and coverage & Prediction, reject, or handoff & Refusal and coverage semantics for the acquisition decision \\
Loop verification & Candidate hypothesis and verification evidence & Accept, reject, or defer & Hypothesis status joined with a separately computed physical transform \\
Graph optimization and sparse computation & Pose graph, topology, and elimination structure & Trajectory estimate and solver cost & Provides the downstream pose-graph and sparse-factor substrate for selected \CARD{} actions \\
\bottomrule
\end{tabularx}
\end{table}
\section{CARD Objects and Causal Interface}
The central distinction is between the object certified by an estimator and the transform executed by a robot. \CARD{} represents them separately and defines the evidence needed to combine them into a service. Throughout the paper, \CARD{} names the interface, LCARD names its lifted transition objective, the \CARD{} controller names its acquisition rule, and a \CARD-selected graph is the intervention topology produced by that rule.

\subsection{Certificate semantics and executable action}
Consider demands indexed by $t=1,\ldots,T$. A request may return no witness or an accepted token $\tau$. The token carries an exact witness assignment $\omega$ for a semantic action $a$. A proof-carrying state and its semantic projection are
\begin{equation}
x=(a,\omega),\qquad \pi(x)=a,
\qquad \mathcal F_t(a)=\{\omega:(a,\omega)\in\mathcal K_t^X\},
\label{eq:lifted-state}
\end{equation}
where $\mathcal K_t^X$ is the legal lifted state set at demand $t$ and $\mathcal F_t(a)$ is the corresponding witness fiber. Its semantic projection is $\mathcal K_t^A=\pi(\mathcal K_t^X)$. The trace begins at a fixed state $x_0=(a_0,\omega_0)$ and selects one state from each $\mathcal K_t^X$.

The certificate-side estimate belongs to the mathematical proof statement. A separate execution map supplies the physical transform $g_t$. The service predicate tests whether that transform has the request-time support required for execution. Checker confirmation and service support then create a proof-backed service. Reference comparison and graph measurements enter the completed evaluation record after the decision.

The causal stream distinguishes an opportunity, an issued request, an accepted token, and the witness assigned to an adopted action. This ordering also separates semantic movement, witness repair, request and storage cost, service cost, latency, and graph-intervention cost. Each quantity is attached to the object that generates it.

\begin{table}[t]
\centering
\caption{CARD notation for proof-carrying states, service, and causal control.}
\label{tab:card-notation}
\footnotesize
\setlength{\tabcolsep}{1.5pt}
\begin{tabularx}{\columnwidth}{@{}l Y@{}}
\toprule
Symbol & Role in the certificate-to-action chain \\
\midrule
$t,\;i$ & chronological demand and eligible-stream indices \\
$a_t$ & semantic action (silence, hold, or adopt) \\
$\omega_t$ & exact witness assignment, including certificate-free coordinates \\
$x_t=(a_t,\omega_t)$ & proof-carrying state \\
$\pi(x_t)=a_t$ & semantic projection \\
$\mathcal K_t^X,\mathcal K_t^A$ & lifted legal set and its semantic projection \\
$\mathcal F_t(a)$ & witness fiber compatible with semantic action $a$ \\
$w,\tau$ & issued request and accepted token, respectively \\
$g_t$ & causal physical relative-transform map \\
$S_{\rm svc}$ & service-support predicate \\
$A_k,L_k$ & request-time actionability and live-proof indicators \\
$W_i,r_i,h_i,c_i$ & witness, request, confirmation, and cumulative-service variables \\
$U_b$ & useful-service yield under request budget $b$ \\
$C(S),\;{\rm WRC}$ & acquisition cost of token set $S$ and witness repair cost \\
$d_{\rm exec},\;d_{\rm int}$ & execution/service and downstream intervention costs \\
$\mathcal F_t^{\rm obs}$ & request-time causal information filtration \\
\bottomrule
\end{tabularx}
\end{table}

Equation~\eqref{eq:lifted-state} separates semantic availability from exact proof support. An empty fiber marks an action with no live witness. A nonempty fiber identifies the witnesses that can support the action, while $g_t$ and $S_{\rm svc}$ define its physical transform and service status.

\subsection{Semantic movement, witness repair, and acquisition}
Let $d_A$ measure semantic movement and $d_W$ measure witness change. For a legal semantic trace $a_{1:T}$, define semantic CARD (SCARD), witness-repair cost (WRC), and lifted CARD (LCARD) together:
\begin{align}
d_X((a,\omega),(a',\omega'))
  &:=d_A(a,a')+d_W(\omega,\omega'), \label{eq:additive-metric}\\
\mathrm{SCARD}
  &:=\min_{a_t\in\mathcal K_t^A}\sum_{t=1}^{T}d_A(a_{t-1},a_t), \label{eq:scard}\\
\mathrm{WRC}(a_{1:T})
  &:=\min_{\omega_t\in\mathcal F_t(a_t)}\sum_{t=1}^{T}d_W(\omega_{t-1},\omega_t), \label{eq:wrc}\\
\mathrm{LCARD}
  &:=\min_{x_t\in\mathcal K_t^X}\sum_{t=1}^{T}d_X(x_{t-1},x_t).
\label{eq:lcard}
\end{align}
Request and storage cost, service cost, latency, and graph-intervention cost are recorded outside this structural metric. The controller observes a filtration $\mathcal F_t^{\rm obs}$ containing current geometry, navigation, support, immutable identities, and completed past outcomes.

\subsubsection{Exact semantic--witness decomposition}
The semantic and witness terms must be evaluated on the same trace. Finite legal fibers and the additive metric in Eq.~\eqref{eq:additive-metric} give the following decomposition.
\begin{theorem}[Lifted fiber decomposition]
\begin{equation}
\mathrm{LCARD}=\min_{a_t\in\mathcal K_t^A}
\left[\sum_{t=1}^{T}d_A(a_{t-1},a_t)+\mathrm{WRC}(a_{1:T})\right].
\label{eq:lifted-decomposition}
\end{equation}
\end{theorem}
Every legal lifted trace belongs to one semantic projection and one compatible witness trace. Minimizing the witness term within each projection cell yields Eq.~\eqref{eq:lifted-decomposition}. Appendix~\ref{app:lifted-proof} gives the full partition argument. The result permits a stable semantic action to carry a changing witness and nonzero repair cost.

\subsubsection{Projection gap and coincidence boundary}
Define the semantic optimum set
\begin{equation}
\mathcal A^\star:=\underset{a_t\in\mathcal K_t^A}{\arg\min}
\sum_{t=1}^{T}d_A(a_{t-1},a_t).
\label{eq:semantic-optimum}
\end{equation}
\begin{proposition}[Projection gap and sharp coincidence]
\begin{equation}
\mathrm{SCARD}\leq\mathrm{LCARD},\qquad
\mathrm{LCARD}=\mathrm{SCARD}
\Longleftrightarrow
\exists a^\star_{1:T}\in\mathcal A^\star:
\mathrm{WRC}(a^\star_{1:T})=0.
\label{eq:projection-coincidence}
\end{equation}
\end{proposition}
Projection removes a nonnegative witness term. Equality requires a semantic-optimal trace with a zero-cost witness path. For the unit identity metric, this is equivalent to
$\{\omega_0\}\cap\bigcap_t\mathcal F_t(a_t^\star)\neq\varnothing$.
A constant action paired with a distinct singleton witness at each demand gives $\mathrm{SCARD}=0$ and $\mathrm{LCARD}=\mathrm{WRC}=T$.

\paragraph{Acquisition consequence.}
Let $S$ be a selected token set, $C(S)$ its acquisition cost, $U(a)$ the obligations of a fixed semantic trace, and $\mathcal F_t^S(a_t)$ the fiber induced by $S$. Acquisition and repair form one optimization:
\begin{equation}
\min_{S\,\mathrm{covers}\,U(a)}
\left[C(S)+
\min_{\omega_t\in\mathcal F_t^S(a_t)}
\sum_{t=1}^{T}d_W(\omega_{t-1},\omega_t)\right].
\label{eq:acquisition-repair}
\end{equation}
The same token set determines coverage and witness compatibility. Pre-proof actionability supplies a causal test of this coupling before certificate purchase.

\paragraph{Service consequence.}
\begin{corollary}[Persistent-support service condition]
If a checker-confirmed token arrives within the request budget and deadline, remains live throughout the service horizon, supports every obligation, and satisfies the transform-coupling constraints, then the detected proposal instantiates a proof-backed service for that demand.
\end{corollary}
Assigning the token to every required witness coordinate gives the construction. Exact-16 correspondences and checker confirmation provide the proof object, causal navigation provides the physical transform, and the service predicate provides request-time eligibility.
\section{System Interface}
\begin{figure}[t]
\centering
\includegraphics[width=\textwidth]{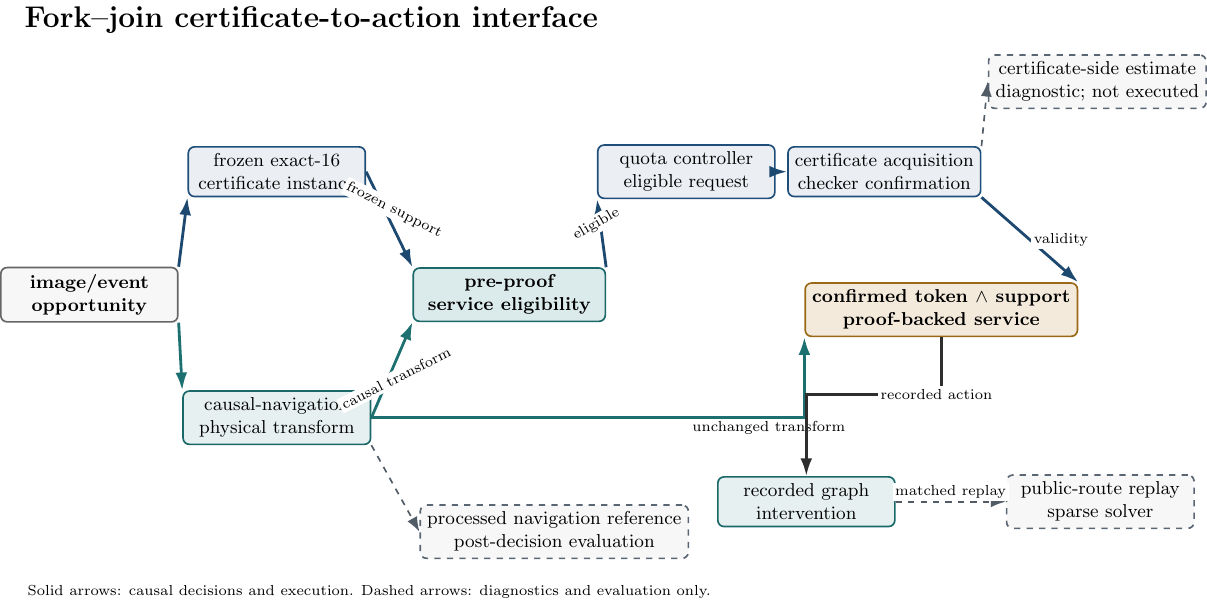}
\caption{The \CARD{} fork--join interface. One image/event opportunity produces a frozen exact-16 certificate instance and a causal-navigation transform. Request-time support determines eligibility, checker confirmation supplies the proof token, and the two channels join at service creation. Dashed paths mark post-decision measurements.}
\label{fig:card-architecture}
\end{figure}

\subsection{Certificate and execution channels}\label{sec:certificate-execution}
Candidate $k$ is an immutable exact-16 instance $\mathcal C_k=\{(a_i,b_i,\beta_i)\}_{i=1}^{16}$. The points $a_i,b_i\in\mathbb R^3$ belong to linked source and target image events, and $\beta_i>0$ is their combined Euclidean proof-support radius. The certificate solves
\begin{equation}
\begin{aligned}
\Phi_k(R,t,s)&=\sum_{i=1}^{16}
\min\!\left\{\frac{\|b_i-(sRa_i+t)\|_2^2}{\beta_i^2},1\right\},\\
\widehat z_k&=(\widehat R_k,\widehat t_k,\widehat s_k)
\in\underset{R\in\mathrm{SO}(3),\,t\in\mathbb R^3,\,s=1}{\arg\min}\ \Phi_k(R,t,s).
\end{aligned}
\label{eq:certificate-tls}
\end{equation}
The scale is fixed to one, so Eq.~\eqref{eq:certificate-tls} is a full-denominator heteroscedastic truncated least-squares registration. The producer enumerates every binary mask, emits the corresponding lower bounds and feasible primal transforms, and reports $L_k$ and $U_k=\Phi_k(\widehat z_k)$. The checker rebuilds all sixteen observations, identities, masks, bounds, and objectives. It confirms the instance when $L_k\leq\inf\Phi_k\leq U_k$, $U_k-L_k\leq10^{-7}$, and the supported subset satisfies the frozen rank conditions. Appendix~\ref{app:geometric-admissibility} records candidate construction and thresholds, while Appendix~\ref{app:measurement-adapter} records the deterministic radius adapter. All instance fields are frozen before optimization.

A confirmed completion creates token $\tau_k$ with immutable identity, support, completion time, validity interval, and checker state. These fields determine its membership in the witness fibers. Token identity remains explicit because the same semantic proposal may be supported by different tokens over time.

\subsection{Certificate-to-action composition}\label{sec:certificate-action-composition}
Let $W$ be the delivered world frame, $C_e$ the provider camera frame at event $e$, $c_e^W$ the camera center, and $R_{WC,e}$ the camera-to-world rotation. The matrix $A=\left[\begin{smallmatrix}0&0&1\\1&0&0\\0&1&0\end{smallmatrix}\right]$ maps action-camera coordinates to the provider optical convention. For the convention $x_t=R_{s,t}x_s+t_{s,t}$, causal navigation computes
\begin{equation}
\begin{aligned}
R_{WC,e}&=R_x(\omega_e)R_y(\phi_e)R_z(\kappa_e),\\
g_{s,t}&=(R_{s,t},t_{s,t}),\\
R_{s,t}&=A^\top R_{WC,t}^\top R_{WC,s}A,\\
t_{s,t}&=A^\top R_{WC,t}^\top(c_s^W-c_t^W).
\end{aligned}
\label{eq:causal-navigation-transform}
\end{equation}
This composition is denoted by $\operatorname{Nav}(I_s,I_t,N_{\leq t})$. The processed navigation reference is converted into the same camera frame using $R^{\rm ref}_{WC,e}=R^{\rm ref}_{WB,e}R_{BC}$ and $c_e^{W,\rm ref}=n_e^{W,\rm ref}+R^{\rm ref}_{WB,e}\ell_C^B$. The fixed camera-to-body rotation $R_{BC}$ and lever $\ell_C^B$ are estimated on nonproposal events. Immutable image/event identities join the causal transform, reference transform, and certificate record.

For the same exact-16 instance, $\chi_{s,t}$ contains the frozen linkage, correspondence identities, and radii. The residual, support count, service predicate, and proof-backed service indicator are
\begin{equation}
\begin{aligned}
\rho_i(g_{s,t})&=\|b_i-(R_{s,t}a_i+t_{s,t})\|_2,\\
n_{\rm sup}(g_{s,t})&=\sum_{i=1}^{16}\mathbf 1\{\rho_i(g_{s,t})\leq\beta_i\},\\
S_{\rm svc}(g_{s,t},\chi_{s,t})
&=\mathbf 1\{n_{\rm sup}(g_{s,t})\geq3\},\\
\operatorname{Service}_{s,t}
&=\mathbf 1\{h_{s,t}=\mathrm{confirmed}\}S_{\rm svc}(g_{s,t},\chi_{s,t}).
\end{aligned}
\label{eq:proof-backed-service}
\end{equation}
The predicate is evaluated before purchase from the causal transform and frozen candidate. Checker confirmation later joins the unchanged predicate to create the service. In this contract, the certificate applies to $\mathcal C_k$ and Eq.~\eqref{eq:certificate-tls}, while $g_{s,t}$ is the execution output.

\subsection{Causal acquisition}
\subsubsection{Pre-proof actionability}
The selector evaluates the causal-navigation transform under $S_{\rm svc}$ before invoking the certificate optimizer. Candidates that can instantiate the service form the eligible stream used by the acquisition controller.

For a frozen population of $N$ candidates, let $A_k\in\{0,1\}$ indicate request-time service eligibility. Let $L_k\in\{0,1\}$ record whether a requested proof completes by its deadline, has a live token at service time, retains a legal witness, and is checker-confirmed. For a budget-$b$ selected set $B_b$, useful yield is
\begin{equation}
U_b=\sum_{k\in B_b} A_kL_k.
\label{eq:useful-yield}
\end{equation}
In the frozen comparison population, $L_k=1$ for every candidate.

\begin{theorem}[Actionability dominance]
Let $K=\sum_{k=1}^{N}A_k$ and assume $L_k=1$ for all $k$. If $B_b^{\rm all}$ is a uniform size-$b$ sample without replacement from all $N$ candidates, then
\begin{equation}
\begin{aligned}
U_b^{\rm all}&\sim\operatorname{Hypergeometric}(N,K,b),\\
\mathbb E[U_b^{\rm all}]&=b\frac{K}{N},\\
\operatorname{Var}(U_b^{\rm all})&=b\frac{K}{N}\left(1-\frac{K}{N}\right)\frac{N-b}{N-1}.
\end{aligned}
\label{eq:hypergeometric-yield}
\end{equation}
If $B_b^{\rm elig}$ samples without replacement from the $K$ eligible candidates and $b\leq K$, then $U_b^{\rm elig}=b$. Its expected advantage over sampling from the full population is $b(1-K/N)$.
\end{theorem}

The two sampling schemes apply the same without-replacement rule to different populations. In the policy block, $N=170$ and $K=155$, so Eq.~\eqref{eq:hypergeometric-yield} gives five parameter-free predictions at the frozen budgets.

\subsubsection{Chronological quota control}
For target $q$, eligible demands are processed in chronological order. Let $W_i$ indicate that a legal witness exists at eligible index $i$, let $h_i$ indicate checker-confirmed completion, and let $c_i$ count confirmed services through index $i$. The synchronous controller is
\begin{equation}
r_i=W_i\mathbf 1\{c_{i-1}<\lceil qi\rceil\},
\qquad c_i=c_{i-1}+r_i h_i,
\qquad c_0=0.
\label{eq:quota-controller}
\end{equation}
When several witnesses are legal, causal robust-cost ordering selects one and immutable identity resolves the final tie. No-witness events do not issue requests, while later confirmed completions advance the remaining quota. The bounded-overlap replay uses the same eligibility and quota equations with at most two requests in flight and timestamp-ordered completions.

\subsection{Downstream graph intervention}
The downstream module maps each selected service to a graph action: delete, downweight, or retain. Its replay interface records immutable edge identity, action label, downweight level, and effective weight. These fields define the selected topology passed to trajectory optimization and native sparse-solver instrumentation. Section~\ref{sec:experimental-evaluation} gives the matched comparison design.
\section{Evidence Design and Evaluation Contract}
\label{sec:experimental-evaluation}
\subsection{Experimental design}
The primary study uses an AA15--X500 airborne survey system equipped with a DG4M2-labeled five-camera oblique-imaging array. Twenty flight sorties contain 6,544 synchronized exposure events and 32,720 five-channel images from mountainous, village, and town scenes. The Full16 evaluation uses synchronized image/event identities, causal navigation fields, and processed navigation products from the same observations.

The evidence architecture combines this proprietary primary study with public downstream evaluation (Table~\ref{tab:evidence-architecture}). The primary study measures the certificate/action boundary, service creation, and chronological request efficiency. The public study measures the trajectory and sparse-solver response of recorded intervention topology on KITTI, EuRoC, and TUM. Each study retains its own experimental unit and denominator.

\begin{table}[!t]
\centering
\caption{Evidence architecture for the primary certificate/action study and public downstream evaluation.}
\label{tab:evidence-architecture}
\footnotesize
\setlength{\tabcolsep}{3pt}
\begin{tabularx}{\textwidth}{@{}P{0.15\textwidth} P{0.24\textwidth} P{0.18\textwidth} Y@{}}
\toprule
Evidence module & Platform and setting & Denominator & Measured quantities \\
\midrule
Primary certificate/action study & AA15--X500 airborne five-camera survey in mountainous, village, and town scenes & 20 sorties, 404 Full16 cases, and 530 policy opportunities & Certificate object, execution transform, service creation, and proof-acquisition yield \\
Public downstream study & KITTI vehicle, EuRoC MAV, and TUM handheld RGB-D routes & 11 matched SLAM cases and 671 instrumented graphs & Trajectory response, sparse fill, numeric solve, and optimizer wall time \\
\bottomrule
\end{tabularx}
\end{table}

The four Full16 blocks are non-overlapping subsets of the airborne acquisition chain. They contain 50 sequential-service certificates, 5 prospective-camera certificates, 179 independent hold-out certificates, and 170 policy-validation certificates. The last block contains 530 chronological opportunities. All four blocks contribute to the 404-certificate transform comparison, while the policy block also supplies the prospective acquisition study.

Each exact-16 candidate freezes sixteen image/event correspondences, positive support radii, and an immutable proposal identity. The certificate optimizer and checker use Eq.~\eqref{eq:certificate-tls}. Causal navigation computes the paired physical transform with Eq.~\eqref{eq:causal-navigation-transform}, and Eq.~\eqref{eq:proof-backed-service} creates the service. The processed navigation reference is attached to the completed record.

The policy study processes opportunities chronologically and records witness supply, repeated actions, replacement, requests, confirmations, latency, and storage cost. A useful service is an executed service supported by the processed-reference annotation. Parameters are fixed at 70\%, 80\%, 90\%, 95\%, and 100\% coverage before reference evaluation. Ordinary matched random and actionability-matched random use the same request budgets, so their difference isolates the pre-proof eligibility screen.

The downstream study covers KITTI keyframe, KITTI frame-dense, EuRoC native-pose, and TUM native-pose routes. For each of 11 cases, ordinary and stratified matched random interventions use 30 fixed seeds per family. The comparisons match delete, downweight, and retain counts, downweight levels, effective total weight, node count, and total edge count. The stratified comparator permutes action labels within each source-vertex stratum. Relative ATE/RPE change measures trajectory response. Optimizer wall time and native sparse-factor stages measure computation.

The solver replay contains 671 graphs and 13,420 native records, with 20 Gauss--Newton iterations per graph. Instrumentation records Hessian structural nonzeros, quadratic-form assembly, symbolic factor size and fill, numeric factorization and triangular solve, and wall time. Graph selection precedes these measurements. Reference transforms, proof outcomes, and solver records are joined after the corresponding decisions.

\begin{table}[!t]
\centering
\caption{Full16 outcomes from four non-overlapping blocks of the AA15--X500 acquisition chain.}
\label{tab:card-block-outcomes}
\small
\setlength{\tabcolsep}{4pt}
\textit{(a) Certificate and physical-transform outcomes}\par\vspace{2pt}
\resizebox{\textwidth}{!}{%
\begin{tabular}{@{}l r r r r r@{}}
\toprule
Block & Certificates & Optimizer discrepant & Mixed & Optimizer consistent & Navigation consistent \\
\midrule
Sequential-service & 50 & 50 & 0 & 0 & 50 \\
Prospective-camera & 5 & 5 & 0 & 0 & 5 \\
Independent hold-out & 179 & 179 & 0 & 0 & 179 \\
Policy-validation & 170 & 170 & 0 & 0 & 170 \\
\midrule
Total & 404 & 404 & 0 & 0 & 404 \\
\bottomrule
\end{tabular}}

\vspace{7pt}\textit{(b) Proof-backed service outcomes}\par\vspace{2pt}
\begin{tabular}{@{}l r r r@{}}
\toprule
Block & Navigation consistent & Proof-backed services & Withheld \\
\midrule
Sequential-service & 50 & 42 & 8 \\
Prospective-camera & 5 & 5 & 0 \\
Independent hold-out & 179 & 168 & 11 \\
Policy-validation & 170 & 155 & 15 \\
\midrule
Total & 404 & 370 & 34 \\
\bottomrule
\end{tabular}
\end{table}

\section{Results}
\subsection{Certificate and execution produce different outcomes}
All 404 exact-16 certificates are checker-confirmed. Their certificate-side optimizer points are reference-discrepant in all 404 cases, with zero mixed and zero consistent annotations. At the same image/event identities, all 404 causal-navigation transforms are reference-consistent. The optimizer/reference median discrepancies are $1.9891042382^\circ$ and $21.6322004178\,\mathrm{m}$, compared with $0.0036566143^\circ$ and $0.0104518618\,\mathrm{m}$ for navigation/reference. The paired result assigns mathematical confirmation to the optimizer coordinate of Eq.~\eqref{eq:certificate-tls} and physical execution to the causal transform of Eq.~\eqref{eq:causal-navigation-transform}.

\begin{figure}[!t]
\centering
\includegraphics[width=\textwidth]{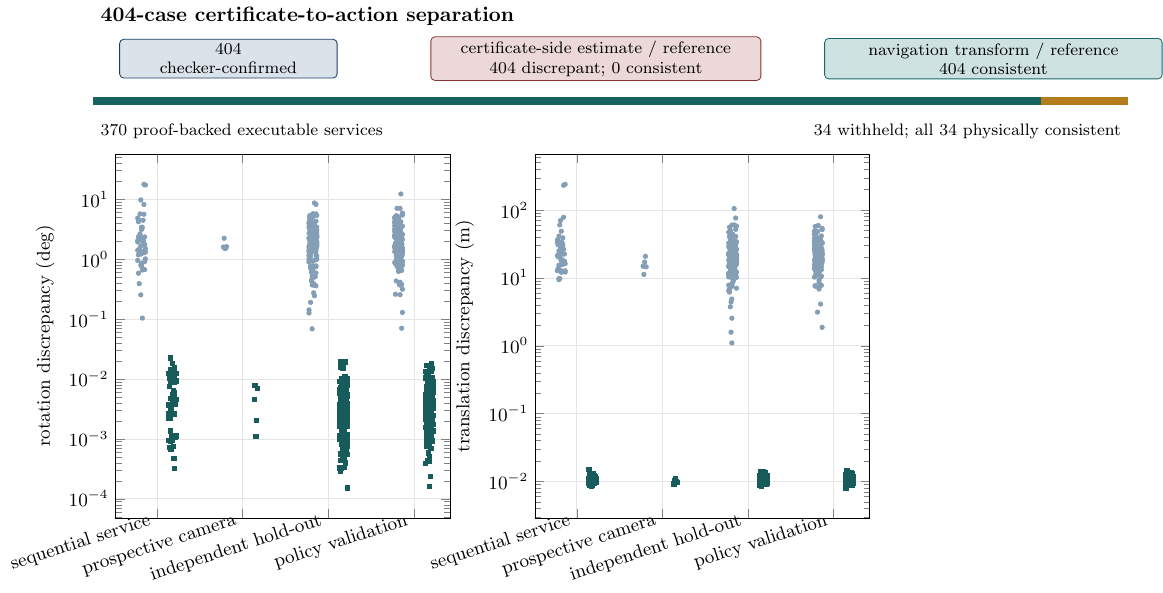}
\caption{Certificate and execution outcomes across 404 cases. Every checker-confirmed optimizer point is reference-discrepant, while every causal-navigation transform is reference-consistent. Optimizer/reference medians are $1.9891^\circ$ and $21.6322\,\mathrm{m}$. Navigation/reference medians are $0.0036566^\circ$ and $0.0104519\,\mathrm{m}$. Block names match Table~\ref{tab:card-block-outcomes}.}
\label{fig:card-discrepancy}
\end{figure}

\subsection{Proof support determines service creation}
Among the 404 navigation-consistent transforms, 370 satisfy the proof-backed service predicate and execute, while 34 are withheld. All 370 executed services are reference-consistent. The 370/34 split measures the support condition that joins a confirmed certificate to its physical transform.

The service stream contains 120 opportunities: 50 witness-available services and 70 no-witness events. It includes 112 unique actions and eight repeated actions. Two actions have simultaneous witnesses, and two later opportunities replace a witness that has become unavailable. The 50 services require approximately 72.55 wall-s, 81.46 CPU-s, 178.8~MiB of proof storage, and 373.3~MiB of request storage. Median service delay is approximately 1.32~s.

\begin{table}[!t]
\centering
\caption{Service-stream realization of the lifted-state objects.}
\label{tab:card-service-stream}
\small
\setlength{\tabcolsep}{5pt}
\begin{tabularx}{\textwidth}{@{}l r Y@{}}
\toprule
Observed object & Count & Lifted-state interpretation \\
\midrule
Chronological opportunities & 120 & Complete service-stream denominator \\
Witness-available / no-witness & 50 / 70 & Nonempty and empty legal witness fibers \\
Unique / repeated actions & 112 / 8 & Semantic identity can recur across demands \\
Actions with simultaneous witnesses & 2 & One action can admit several witnesses \\
Later replacement opportunities & 2 & Witness identity changes over time \\
\bottomrule
\end{tabularx}
\end{table}

\subsection{Pre-proof eligibility determines useful-service yield}
The request-time support predicate admits 155 of the 170 exact-16 policy candidates. All 170 candidates are checker-confirmed in the completed record, and all 170 navigation transforms are reference-consistent. The eligible population supplies 155 proof-backed services.

At 70\%, 80\%, 90\%, 95\%, and 100\% useful-service targets, \CARD{} requests 109, 124, 140, 148, and 155 certificates and obtains the same number of useful services. Equation~\eqref{eq:hypergeometric-yield} predicts ordinary-random means of 99.3824, 113.0588, 127.6471, 134.9412, and 141.3235 services. The observed ten-seed means are $99.8\pm1.60$, $113.9\pm1.14$, $127.1\pm0.70$, $135.2\pm1.99$, and $141.2\pm0.75$. The largest prediction residual is 0.8412 services. At equal request budgets, \CARD{} gains 9.2, 10.1, 12.9, 12.8, and 13.8 useful services over the observed ordinary-random means. Table~\ref{tab:card-frontier-cost} includes the quota-only and cost/expected-confirmation selectors.

The bounded-overlap completion-time replay selects the same request identities and the same five request/useful vectors as synchronous \CARD{}. It admits at most two concurrent requests and processes completions in timestamp order. Request identity is unchanged at every target, and the stale-confirmed event at each target adds no request.

\begin{figure}[t]
\centering
\includegraphics[width=0.96\textwidth]{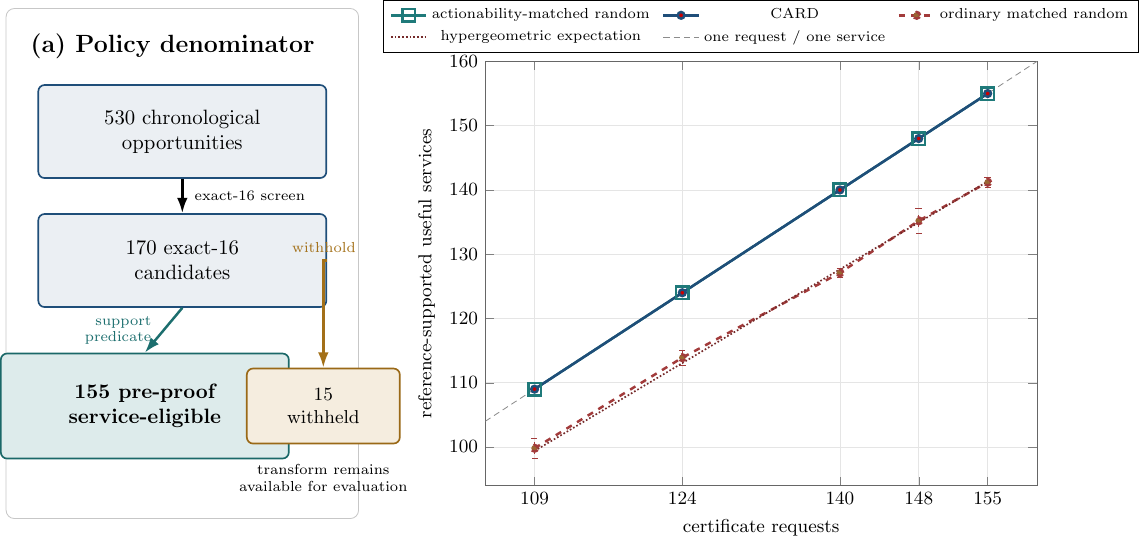}
\caption{Policy denominator and useful-service frontier. The causal screen maps 530 opportunities to 170 Full16 candidates and 155 pre-proof service-eligible candidates. \CARD{} and actionability-matched random lie on the one-request/one-useful-service line at all five targets. Ordinary-random means follow the Hypergeometric$(170,155,b)$ expectation. Error bars show one standard deviation across ten seeds.}
\label{fig:card-frontier}
\end{figure}

\begin{table}[!b]
\centering
\caption{Useful-service frontier and resource savings relative to always-buy. The hypergeometric column is fixed by $N=170$, $K=155$, and request budget. Ordinary-random entries report mean $\pm$ standard deviation across ten seeds.}
\label{tab:card-frontier-cost}
\small
\setlength{\tabcolsep}{3.2pt}
\textit{(a) Useful-service frontier}\par\vspace{2pt}
\resizebox{\textwidth}{!}{%
\begin{tabular}{@{}c r r r r r r@{}}
\toprule
Target & \CARD{} req/use & Ordinary theory & Ordinary observed & Quota-only req/use & Cost/expected req/use & Always-buy req/use \\
\midrule
70\% & 109/109 & 99.3824 & $99.8\pm1.60$ & 131/119 & 156/143 & 170/155 \\
80\% & 124/124 & 113.0588 & $113.9\pm1.14$ & 149/136 & 160/146 & 170/155 \\
90\% & 140/140 & 127.6471 & $127.1\pm0.70$ & 167/153 & 164/150 & 170/155 \\
95\% & 148/148 & 134.9412 & $135.2\pm1.99$ & 170/155 & 166/152 & 170/155 \\
100\% & 155/155 & 141.3235 & $141.2\pm0.75$ & 170/155 & 170/155 & 170/155 \\
\bottomrule
\end{tabular}}

\vspace{7pt}\textit{(b) Resource savings versus always-buy}\par\vspace{2pt}
\begin{tabular}{@{}c r r r r@{}}
\toprule
Target & Requests saved & Wall saved (s) & Proof saved (MiB) & Request storage saved (MiB) \\
\midrule
70\% & 61 & 89.200 & 215.9 & 450.9 \\
80\% & 46 & 59.655 & 141.8 & 297.7 \\
90\% & 30 & 37.573 & 88.2 & 185.4 \\
95\% & 22 & 31.425 & 75.2 & 157.2 \\
100\% & 15 & 18.267 & 41.8 & 88.2 \\
\bottomrule
\end{tabular}
\end{table}

\FloatBarrier
\subsection{Selected topology changes sparse computation}
The matched graph design fixes delete, downweight, and retain counts, downweight levels, effective total weight, node count, and total edge count across 11 cases. Stratified matched random also preserves these quantities within each source-vertex stratum.

\begin{figure}[!t]
\centering
\includegraphics[width=0.98\textwidth]{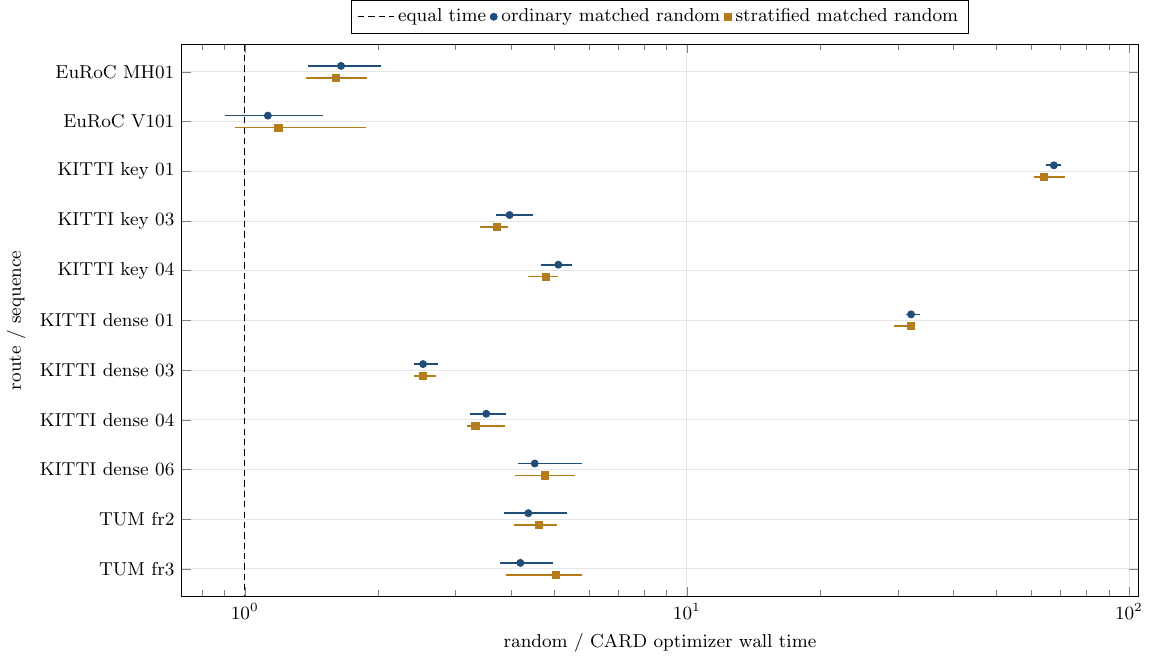}
\caption{Matched real-SLAM optimizer runtime across 11 cases. Both random families lie above the equal-time line in every case. Markers show median random/\CARD{} ratios, and segments show the central 90\% of 30-seed distributions. Maximum absolute relative ATE/RPE change is 0.0611\%.}
\label{fig:card-runtime}
\end{figure}

The maximum absolute relative ATE/RPE change is 0.0611\%, and \CARD{} retains 1,524--55,005 covisibility edges. Ordinary random is $1.128\times$--$67.65\times$ slower in optimizer wall time across the 11 cases. Both random families are slower in every case (Fig.~\ref{fig:card-runtime}). With graph size, intervention budget, and effective weight fixed, the comparison assigns the measured computation difference to selected topology.

\begin{table}[!b]
\centering
\caption{Matched downstream design and measured runtime and sparse-solver outcomes.}
\label{tab:card-downstream}
\small
\begin{tabularx}{\textwidth}{@{}l l Y l@{}}
\toprule
Comparison & Denominator & Matched quantity or outcome & Pattern \\
\midrule
Graph intervention & 11 cases $\times$ 60 random graphs & Actions, levels, effective weight, nodes, and total edges & All seven quantities matched \\
Trajectory response & 11 cases & Maximum absolute relative ATE/RPE change = 0.0611\% & Small and mixed \\
Retained structure & 11 cases & \CARD{} retains 1,524--55,005 covisibility edges & Context retained \\
Original timing & 11 cases, 30 seeds per family & Random/\CARD wall time = 1.128$\times$--67.65$\times$ & Random topologies slower \\
Mechanism replay & 671 graphs, 13,420 native records & 20 Gauss--Newton iterations per graph & Native stages localize effect \\
\bottomrule
\end{tabularx}
\end{table}

\FloatBarrier
\subsubsection{Sparse-solver mechanism}
The runtime gap appears after symbolic elimination. Random/\CARD{} Hessian structural nonzeros remain near matched at $0.98583\times$--$1.0029\times$, and quadratic-form assembly remains at $0.95452\times$--$1.0769\times$. The separation begins in the factors. Cholesky factor nonzeros span $1.0549\times$--$6.5328\times$, fill spans $1.0695\times$--$6.5288\times$, and numeric factorization plus triangular solve spans $1.1524\times$--$94.517\times$.

Approximate minimum-degree ordering selects the elimination sequence for sparse normal equations~\cite{amestoy1996amd}. Eliminating a variable connects remaining neighbors and creates entries absent from the original Hessian~\cite{rose1976vertexelimination}. Fill and Cholesky factor size capture topology-dependent symbolic work that raw edge count misses. Within sequences, fill and numeric factorization/solve have median Spearman correlation $\rho=0.9628$. Numeric work and wall time have median $\rho=0.9051$. Both correlations are positive in all 11 sequences.

\CARD{} ranks first for factor nonzeros and wall time among \CARD{} plus 30 corresponding random graphs in all 22 sequence/family comparisons. No random graph matches or beats \CARD{} within its comparison. The measured mechanism runs from selected topology through AMD fill and Cholesky factor size to numeric solve burden and wall time. Complete-graph fill under the native ordering is the measured mediator. Long-range and low-covisibility edge counts provide complementary topology descriptors.

\begin{figure}[!t]
\centering
\includegraphics[width=0.98\textwidth]{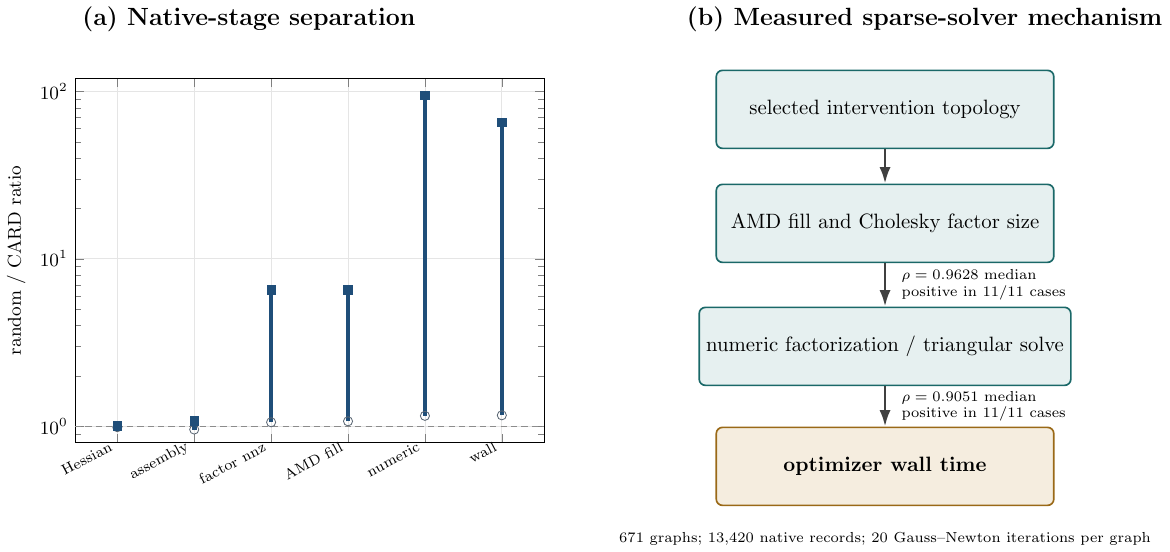}
\caption{Native sparse-solver measurements under matched graph intervention. Hessian structure and assembly remain near one, while factor size, AMD fill, numeric work, and wall time separate. Within-sequence Spearman correlations are positive in all 11 cases. Median $\rho$ is 0.9628 for fill versus numeric work and 0.9051 for numeric work versus wall time.}
\label{fig:card-solver}
\end{figure}

\section{Discussion}
\subsection{Certificate semantics and execution}
Separating certification from execution changes the unit of system design. The 404-case comparison attaches checker confirmation and processed-reference consistency to different outputs. The optimizer point belongs to the global-bound statement for a frozen Full16 registration instance, while causal navigation supplies the physical transform. Their shared image/event identity makes this boundary directly measurable.

The lifted-fiber result extends the distinction to service time. A stable semantic relation can coexist with changing witness identity, empty fibers, or replacement. The measured service stream contains each of these states. The certified object, execution map, and witness-support contract together define the service.

This formulation complements certifiable estimation and robust registration by assigning each guarantee to its system object. The processed reference evaluates the completed physical transform, while the certificate and controller histories retain their original identities. Mathematical validity and physical consistency can then be measured independently on the same candidate.

\subsection{Actionability and service yield}
Pre-proof eligibility tests whether a certificate purchase can instantiate the required service. Ordinary matched random spends the same budgets on a population containing 15 service-ineligible candidates. Actionability-matched random restores the eligible sampling population and reproduces \CARD{}'s yield. Because both controls use stable uniform ranks and prefix selection, the comparison localizes the gain to eligibility rather than chronological scheduling.

With $N=170$ and $K=155$, ordinary matched random has expected yield $155b/170$, while actionability-matched acquisition yields $b$ under the observed live-confirmation condition. The five ordinary-random means remain within $0.8412$ services of these parameter-free predictions. The measured frontier follows the population change created by screening service eligibility before proof purchase.

Quota control operates after this screen on a stream whose witnesses arrive, expire, repeat, or require replacement. Service semantics determine which proof requests can be useful. The chronological controller determines when those requests are needed.

\subsection{Topology and sparse computation}
Selected topology changes computation through elimination structure. Matching graph size, intervention counts, effective weight, and Gauss--Newton iterations leaves AMD fill and Cholesky factor size free to vary. Native measurements locate the runtime separation at these symbolic and numeric stages, after Hessian structure and assembly~\cite{kuemmerle2011g2o,dellaert2017factorgraphs}.

The mechanism connects recorded action selection to backend cost. \CARD{} retains substantial covisibility structure and produces smaller factors under the native ordering, with a maximum absolute relative ATE/RPE change of 0.0611\%. Edge count and effective weight do not capture this elimination structure. Fill of the complete selected topology is the measured mediator.

Near-matched Hessian structural nonzeros place the difference after graph assembly. Equal node, edge, intervention, and effective-weight budgets hold graph size and total weight fixed. Positive fill-to-numeric and numeric-to-wall correlations in all 11 sequences connect the symbolic change to optimizer time.

\subsection{Proprietary primary study and public downstream evaluation}
The empirical architecture combines depth in a proprietary primary system with reproducibility in public downstream graphs. The AA15--X500 study measures the certificate/action/service boundary across 404 non-overlapping cases from 20 sorties and mixed aerial-survey scenes. KITTI, EuRoC, and TUM measure trajectory and sparse-solver responses on vehicle, aerial, and handheld SLAM routes.

The synchronous and bounded-overlap controllers preserve identical selected requests and useful-service counts at every frontier target. The public-route module carries the recorded intervention labels into matched graph replay, where native instrumentation measures trajectory response, symbolic fill, numeric factorization, and wall time.

The two modules share the \CARD{} service and intervention objects. Module-specific denominators keep the physical-transform, acquisition, and computational measurements distinct.
\FloatBarrier
\section{Conclusion}
\CARD{} connects a frozen certificate objective, a causal physical transform, request-time service support, proof acquisition, and recorded graph intervention. The lifted-fiber decomposition identifies the witness cost hidden by semantic projection, and the actionability theorem predicts useful-service yield at a fixed request budget. In the primary study, all 404 checker-confirmed optimizer points are reference-discrepant while all 404 causal-navigation transforms are reference-consistent. The five policy means follow the hypergeometric prediction, and matched acquisition recovers one useful service per request. Public downstream comparisons trace the optimizer-time difference to AMD fill, Cholesky factor size, and numeric work. Keeping the certified object, physical transform, service predicate, acquisition rule, and graph action explicit turns mathematical evidence into a measurable system interface.

\appendix
\section{Proof sketches and supplementary definitions}
The canonical model starts with a fixed initial proof-carrying state $(a_0,\omega_0)$ and finite legal sets $\mathcal K_t^X\subseteq\mathcal A\times\Omega$. An accepted token is live at time $t$ only after its declared completion, inside its effective validity interval, and before invalidation. Certificate-free semantic modes use the distinguished witness $\bot$. These conventions make simultaneous completion, expiry, rejection, and invalidation part of legality rather than tunable penalties.

\subsection{Geometric admissibility and exact-16 selection}\label{app:geometric-admissibility}
The frozen admissibility test requires at least 16 joint correspondences and numerical rank of at least two for the source, target, and cross terms. The source and target covariance eigenvalue ratios and the cross-covariance singular-value ratio must be at least $10^{-3}$. Source and target diameter must be at least $10^{-3}\,\mathrm m$, and median pairwise separation divided by diameter must be at least $10^{-2}$. The normalized Horn leading eigengap must be at least $10^{-3}$, while its rotation-information condition number must be at most $10^3$. Numerical rank uses the largest singular value times $\max(n,3)$ times double-precision machine epsilon. The selected exact-16 source and target sets also require minimum pairwise separation divided by diameter of at least $10^{-4}$. Centered coordinates are normalized independently by their RMS radii and concatenated into six-dimensional vectors. Selection starts at the point farthest from the joint centroid and repeatedly maximizes distance to the selected set. Immutable identity hashes resolve exact ties.

\subsection{Measurement adapter reproducibility note}\label{app:measurement-adapter}
The export path feeds the certificate interface through a deterministic registered-RGB-D adapter. Its output is the frozen camera-optical point and combined Euclidean proof-support radius used in Section~\ref{sec:certificate-execution}. The adapter uses $f_x=f_y=1000$ pixels, zero principal point, and a fixed $256\,\mathrm m$ optical-depth offset before backprojection. Outward interval backprojection expands the depth interval until the endpoint radius equals $2\lVert p\rVert_2\sin(1^\circ)$, with outward rounding. Source and target contributions combine as $\beta_i=\beta_i^{\rm src}+\beta_i^{\rm tgt}$ because the scale upper bound is one. The resulting $\beta_i$ is the truncation and proof-support scale in Eq.~\eqref{eq:certificate-tls}.

\subsection{Lifted-fiber decomposition}\label{app:lifted-proof}
The lifted-fiber decomposition follows by grouping every legal proof-carrying trace by its semantic projection. For a fixed semantic trace $a_{1:T}$, every compatible lifted trace has one witness $\omega_t\in\mathcal F_t(a_t)$, and every such witness trace reconstructs one lifted trace. The witness term inside one projection cell is
\begin{equation}
{\rm WRC}(a_{1:T})=
\min_{\omega_t\in\mathcal F_t(a_t)}
\sum_{t=1}^{T} d_W(\omega_{t-1},\omega_t).
\end{equation}
Adding the semantic movement term and minimizing over semantic traces gives the main theorem. The fiber construction accommodates shared tokens, nonproduct legality, and coupled witness assignments. Empty fibers exclude the corresponding semantic trace. An empty lifted legal set gives infeasibility.

\subsection{Projection inequality and linear gap}
Projecting any lifted trace removes only the nonnegative witness summand, yielding $\mathrm{SCARD}\leq\mathrm{LCARD}$. For the linear-gap family, use one constant semantic action, an initial token $\omega_0$ valid only at time zero, and a distinct singleton-lifetime token $\omega_t$ at each later time. The semantic trace has zero movement. Every compatible lifted trace changes witness identity at every step, so under the unit identity metric $\mathrm{WRC}=T$ and $\mathrm{LCARD}=T$. The same construction forces one producing request per time step, showing that semantic stability can hide both repair and acquisition growth.

\begin{table}[t]
\centering
\caption{Toy projection-gap construction. The semantic action is constant, but each successive legal fiber contains a new exact witness.}
\label{tab:toy-projection-gap}
\footnotesize
\setlength{\tabcolsep}{3pt}
\begin{tabular}{@{}c c c c@{}}
\toprule
Time & Semantic action & Live witness fiber & Unit repair \\
\midrule
$0$ & $a$ & $\{\omega_0\}$ & -- \\
$1$ & $a$ & $\{\omega_1\}$ & $1$ \\
$2$ & $a$ & $\{\omega_2\}$ & $1$ \\
$\vdots$ & $a$ & $\{\omega_t\}$ & $1$ per step \\
\bottomrule
\end{tabular}
\end{table}

\subsection{Coincidence and acquisition--repair coupling}
If one accepted token supports every required component/proposal pair, remains live for the horizon, satisfies coupling, and arrives before the first obligation, assigning it to all required coordinates yields zero post-realization repair. This proves the persistent-support service corollary.

For a selected token set $S$, partition feasible acquisition/witness pairs by $S$. Within one partition, the request cost is $C(S)$ and the compatible witness paths are exactly the fibers $\mathcal F_t^S(a_t)$. Minimizing within the partition and then over covering sets yields the joint acquisition--repair proposition. A cover selected without its induced fibers can combine an inexpensive request set with a witness path that those requests cannot realize. This is the formal reason the system evaluates service support before proof purchase.

\subsection{Causal quota argument}
The chronological controller's state contains the current eligible index, confirmed-service count, current legal witness set, completed past outcomes, and immutable identities. A no-witness event leaves the request count unchanged. An unconfirmed completion leaves the service count unchanged. The quota condition is evaluated on the eligible stream, so its request index is distinct from the raw opportunity index.
{\small
\bibliographystyle{unsrt}
\bibliography{refs}
}
\end{document}